%% file: main.tex
\documentclass{article}
\usepackage[utf8]{inputenc}
\usepackage[margin=1.25in]{geometry}

\usepackage{authblk}
\usepackage{graphicx}
\usepackage{makecell}
\usepackage[colorlinks=true,linkcolor=black,citecolor=blue,allbordercolors={1 1 1}]{hyperref}
\usepackage{xspace}

\newcommand{\dalys}{\textsc{Daly}s\xspace}

\newcommand{\hbe}{\textsc{Hbe}\xspace}
\newcommand{\hbes}{\textsc{Hbe}s\xspace}
\newcommand{\hae}{\textsc{Hae}\xspace}
\newcommand{\haes}{\textsc{Hae}s\xspace}

\newcommand{\gps}{\textsc{Gps}\xspace}

\newcommand{\fc}{\textsc{Fc}\xspace}
\newcommand{\cnn}{\textsc{Cnn}\xspace}
\newcommand{\lstm}{\textsc{Lstm}\xspace}

\newcommand{\pr}{\textsc{Pr}\xspace}
\newcommand{\auc}{\textsc{Auc}\xspace}
\newcommand{\roc}{\textsc{Roc}\xspace}
\newcommand{\prauc}{\textsc{Pr-auc}\xspace}
\newcommand{\praucs}{\textsc{Pr-auc}s\xspace}
\newcommand{\rocauc}{\textsc{Roc-auc}\xspace}
\newcommand{\tpr}{\textsc{Tpr}\xspace}
\newcommand{\fpr}{\textsc{Fpr}\xspace}

\title{Smartphone-based Hard-braking Event \\ Detection at Scale for Road Safety Services}
\author[1]{Luyang Liu\thanks{Correspondence to Luyang Liu \{\url{luyangliu@google.com}\} and Shawn O'Banion \{\url{obanion@google.com}\}}}
\author[1]{David Racz}
\author[1]{Kara Vaillancourt}
\author[1]{Julie Michelman}
\author[1]{Matt Barnes}
\author[1]{Stefan Mellem}
\author[1]{Paul Eastham}
\author[1]{Bradley Green}
\author[1]{Charles Armstrong}
\author[1]{Rishi Bal}
\author[1]{Shawn O'Banion}
\author[1,2]{Feng Guo}
\affil[1]{Google Research, Mountain View, CA, USA}
\affil[2]{Department of Statistics, Virginia Tech, Blacksburg, VA, USA}
\date{}

\begin{document}

\maketitle

\input{sections/abstract}
\input{sections/introduction}
\input{sections/data}
\input{sections/methodology}

\input{sections/result}
\input{sections/conclusion}
\input{sections/acknowledgement}




\bibliographystyle{unsrt}
\bibliography{references}
\end{document}

%% file: sections/abstract.tex
\begin{abstract}
Road crashes are the sixth leading cause of lost disability-adjusted life-years (\dalys) worldwide. One major challenge in traffic safety research is the sparsity of crashes, which makes it difficult to achieve a fine-grain understanding of crash causations and predict future crash risk in a timely manner. Hard-braking events have been widely used as a safety surrogate due to their relatively high prevalence and ease of detection with embedded vehicle sensors. As an alternative to using sensors fixed in vehicles, this paper presents a scalable approach for detecting hard-braking events using the kinematics data collected from smartphone sensors. We train a Transformer-based machine learning model for hard-braking event detection using concurrent sensor readings from smartphones and vehicle sensors from drivers who connect their phone to the vehicle while navigating in Google Maps. The detection model shows superior performance with a $0.83$ Area under the Precision-Recall Curve (\prauc), which is $3.8\times$better than a \gps speed-based heuristic model, and $166.6\times$better than an accelerometer-based heuristic model. The detected hard-braking events are strongly correlated with crashes from publicly available  datasets, supporting their use as a safety surrogate. In addition, we conduct model fairness and selection bias evaluation to ensure that the safety benefits are equally shared. The developed methodology can benefit many safety applications such as identifying safety hot spots at road network level, evaluating the safety of new user interfaces, as well as using routing to improve traffic safety.
\end{abstract}

%% file: sections/introduction.tex
\section{Introduction}

Road crashes are the sixth leading cause of lost disability-adjusted life-years (\dalys) worldwide, and the only non-disease cause in the top 15~\cite{leading-causes-of-dalys-who}. As indicated by a \textsc{Who} report~\cite{road-traffic-injuries-who}, approximately 1.3 million people die each year because of road traffic crashes -- higher than other major causes such as \textsc{Hiv}/\textsc{Aids}, tuberculosis or diarrhoeal diseases~\cite{the-top-10-causes-of-death-who}.  Traffic crashes are currently the leading cause of death for children and young adults aged 5–29 years~\cite{road-traffic-injuries-who}. Safety research and applications in the past several decades have led to considerable improvement in road safety. However, to achieve ambitious future goals, e.g.,  Vision Zero~\cite{vision-zero} and US \textsc{Dot}'s strategic plans to reduce fatality by 75\% in the next 30 years~\cite{zero-trafic-death}, novel approaches including new data sources and methodologies are imperative. 

Road safety data sources are essential for understanding crash causation and predicting future risk, both of which are necessary to effectively reduce road traffic crashes. Crashes are affected by many risk factors, such as driving speed, road infrastructure, vehicle factors, traffic conditions, and driver behavior factors.  Many of these factors are time-variant and have a transient effect on crashes, which makes accurately capturing such information a challenging task. Substantial efforts have been made to collect crash information to support safety research.  For example, the National Automotive Sampling System General Estimates System~\cite{nass-general-estimates-system} uses a systematic sampling approach to estimate the overall crash statistics at US level; the Fatality Analysis Reporting System (\textsc{Fars})~\cite{fars} is a census of all fatal crashes in US; the Highway Safety Information System (\textsc{Hsis})~\cite{hssi} contains crash, road infrastructure, and traffic volume information from multiple states. Specialized data, such as the National Motor Vehicle Crash Causation Survey (\textsc{Nmvccs})~\cite{nmvccs} and the \textsc{Shrp{\small 2}} Naturalistic Driving Study~\cite{shrp-2} provide unique information on crash causation and risk.  

\begin{figure}[ht]
    \centering
    \includegraphics[width=01.0\textwidth]{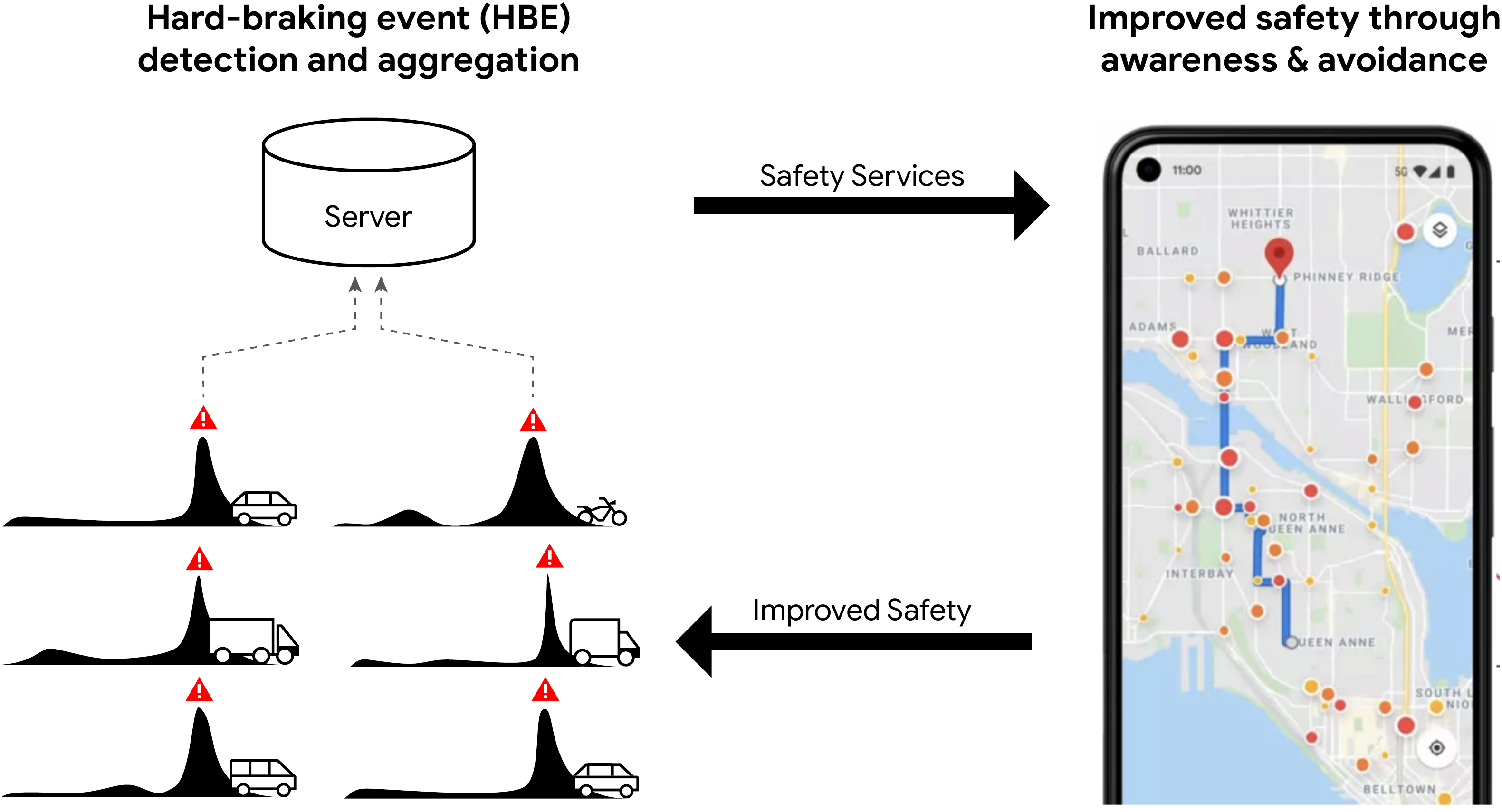}
    \caption{Server aggregates hard-braking events from end devices, and provides safety services back to them.}
    \label{fig:intro_fig}
\end{figure}

While these data provide valuable information, the sparsity of crashes makes it difficult to answer questions such as: (1) How can we detect emergent road conditions in real-time rather than in days? (2) How can we understand road safety in low-traffic road segments with few crashes? (3) How can we route vehicles to safer routes to improve safety based on real-time safety conditions of alternative routes? Answering these questions requires a scalable and rapid way to assess safety at the road network level with high coverage.

To overcome the sparsity of crashes, crash surrogates have been used extensively in road safety research~\cite{gettman2008surrogate}. Crash surrogates are non-crash events that represent crash proneness and are associated with crash risk. For example, the traffic conflicts method was proposed in the 1960s~\cite{perkins1968traffic} and was later adopted by the Federal Highway Administration~\cite{parker1989traffic}. A number of alternative quantitative surrogate measures have also been proposed, such as time to collision~\cite{nadimi2020evaluation}, post-encroachment time~\cite{peesapati2018can}, extended delta-\textsc{v}~\cite{laureshyn2017search}, and near-misses~\cite{samara2020video}. Many of these metrics can be automatically measured using video analytics techniques but many are typically limited to fixed locations. A common premise of surrogates is that they occur at higher frequency and represent abnormal, unsafe conditions, which enables crash risk prediction or assessment without relying on sparse crashes. 
 
One major advancement in the last two decades is the availability of high resolution, in-vehicle collected driving data represented by large-scale naturalistic driving studies (\textsc{Nds})~\cite{dingus2015naturalistic, guo2019statistical}. The objectively collected, continuous driving data not only provides detailed information on driver, vehicle, traffic, and environment moments before crashes but also provides massive amounts of non-crash driving data. The \textsc{Nds} have provided conclusive evidence on the safety risk associated with driver distraction, smartphone use, driver emotional status, age, senior driver, fitness-to-drive, and teenage driver risk~\cite{dingus2016driver, guo2015older, guo2017effects, klauer2014distracted}. Due to the high cost associated with \textsc{Nds} data collection and the rarity of crashes, surrogate measures have been commonly used for crash risk prediction and risk factor identification~\cite{guo2013individual, klauer2014distracted}. Guo et al~\cite{guo2010near} proposed a systematic framework to assess the validity of crash surrogates. A key point is that the validity of surrogate measures depends on the objective of a study, e.g., surrogates for assessing the risk associated with distraction might be different from predicting high risk drivers. 	
			
High G-force events are instances when the acceleration of a vehicle is larger than a predetermined threshold, and are usually caused by hard-braking, swerving, and abrupt acceleration maneuvers. Compared to other surrogates, these high G-force events typically occur at much higher frequency and are relatively easy to identify with embedded vehicle sensors. As such, this form of surrogate has been widely used in crash identification~\cite{hankey2016description}, and evaluation of driver behavior~\cite{simons2009hard, simons2013trajectories}. Research suggests that high G-force events are predictive of crash risk~\cite{simons2012elevated, kim2016exploring, palat2019evaluating}. High G-force events have also been used in teenage risk behavior intervention~\cite{klauer2017driver}. Mao et al~\cite{mao2021decision} investigated the optimal threshold to define a high G-force event and to predict high risk drivers. The literature in general supports the validity of high G-force events as a surrogate, with hard-braking events (large longitudinal deceleration) being one of the most important metrics.  

The ubiquitousness of smartphones offers the opportunity to detect hard-braking events (\hbes) at an unprecedented global scale. \hbes are characterized by a large acceleration in the opposite direction of vehicle longitudinal movement, typically caused by evasive maneuver and sometimes impact with an object. As shown in Figure~\ref{fig:intro_fig}, aggregating \hbes on a central server allows us to understand road safety and provide safety services back to end devices.  These \hbes, along with other context information including location, time, and phone usage, can help provide insights for a variety of safety applications. For example, high granularity \hbe prediction enables a safety-aware routing platform to route drivers, cyclists, and pedestrians away from \hbe-prone areas. \hbe detection at the road network level can support emergent safety hotspot identification in real time. In addition, \hbes can be used to assess the safety impact of specific phone features. Compared to \hbes based on dedicated third party sensors or embedded vehicle sensors, phone sensor-based \hbes provide a cost-effective, always-connected, and scalable solution for the entire road network.   

Accurately detecting \hbes using smartphone sensors is a difficult task. Both accelerometers and \gps speeds are potentially noisy in real world settings. For example, dropping one’s phone on the floor of the vehicle will register a large acceleration signal and may cause a false positive \hbe. Additionally, constant changes in phone orientation make it difficult to assert the direction of the vehicle movement. Most existing research relies either on embedded vehicle sensors~\cite{hull2006cartel} or special equipment to align the smartphone to the vehicle's coordinate system~\cite{mohan2008nericell, thiagarajan2009vtrack, wang2015determining, chen2015invisible, liu2017bigroad}. These pose significant challenges in scaling to more than a few hundred or thousand vehicles. An accurate \hbe detector must be able to distinguish legitimate vehicle acceleration events from other much more common non-\hbe related phone movements.

In this work, we use concurrent smartphone and vehicle sensor readings from Google Maps~\cite{google-maps} navigation drivers to learn a machine learning-based hard-braking event detection model. We train a Transformer-based model~\cite{vaswani2017attention} using phone sensors as features and labels derived from concurrent wheel speed. Optimal models are identified through a comprehensive architecture and hyperparameter search. The Transformer-based models are compared with heuristics-based  algorithms using phone \gps and accelerometer signals, respectively. A correlation analysis is conducted to assess the relationship between phone \hbes and collisions collected from crash databases. In addition, we conduct model fairness and selection bias evaluation to ensure that the safety benefits are equally shared. We conclude with a discussion on the potential usage of this model to improve road safety.

%% file: sections/data.tex
\section{Data}

Several datasets are used in this study: (1) Data from Android Auto projection mode for model training and evaluation; and (2) Data from regular phone use (no vehicle sensor data) and (3) Collision data for evaluating the association between \hbes and crashes.

The projection mode dataset contains concurrent smartphone and vehicle sensor readings from Android Auto users~\cite{android-auto} when they project their phone to the vehicle screen while navigating in Google Maps. A user in the projection mode tethers their phone to a vehicle equipped with an Android Auto system and projects the phone content onto the embedded screen in the vehicle. In projection mode, both smartphone sensor data (including Accelerometer, Gyroscope, \gps speed), and vehicle sensor readings (wheel speed) from the vehicle Controller Area Network are available. The data from vehicle sensors, i.e., the wheel speed data, directly reflects the vehicle kinematics. The \hbes identified from the wheel speed are treated as ground truth about whether—and to what extent—a vehicle is braking.

The availability of concurrent phone and vehicle sensor data in projection mode provides an opportunity to train models to discern actual deceleration events using smartphone sensors only. Specifically, a supervised \hbe detection model is developed using only smartphone sensors as features, and information from vehicle wheel speed as labels. While the data in projection mode only accounts for a relatively small subset of the overall data,  the trained \hbe detection model using phone sensors unlocks the potential for \hbe detection for all smartphone users.

\begin{figure}[ht]
    \centering
    \includegraphics[width=.8\textwidth]{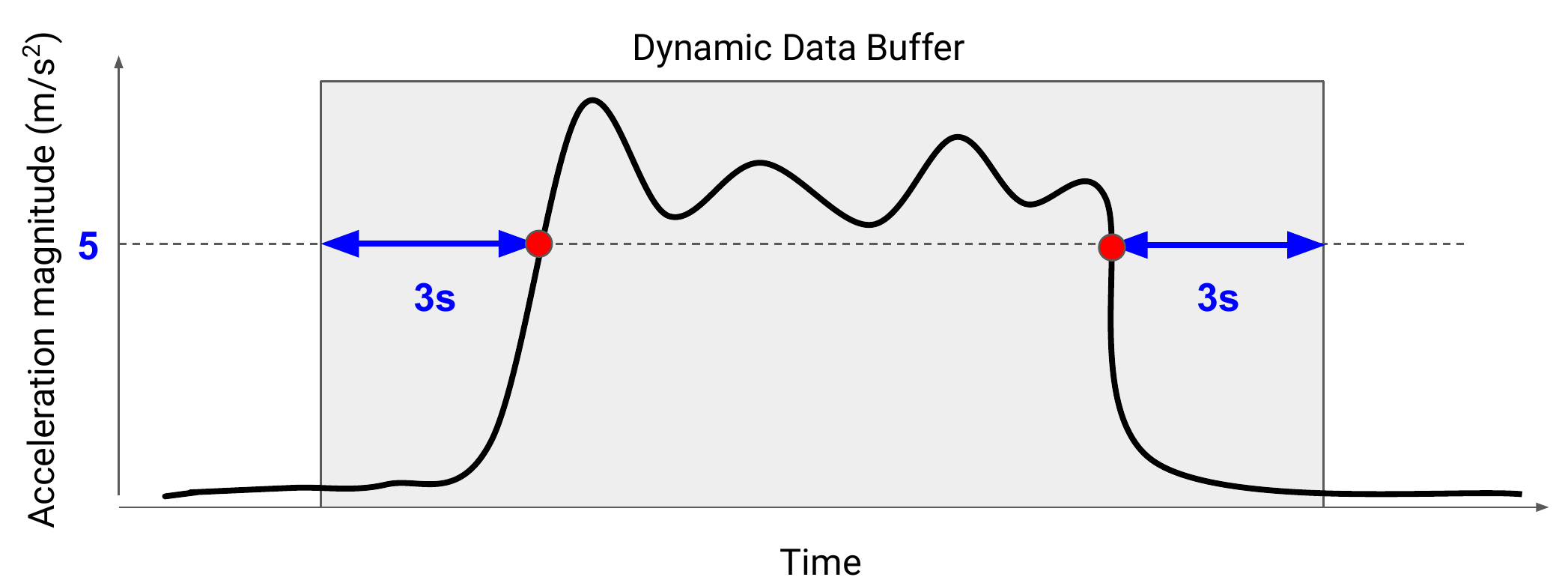}
    \caption{Sliding window based High Acceleration Events (\hae) detection.}
    \label{fig:dynamic_data_buffer}
\end{figure}

Data collection is based on High Acceleration Event (\hae) triggers. \haes are characterized by large linear acceleration readings in any direction, which can be detected by thresholding on the magnitude of the smartphone's linear acceleration readings. We design a sliding window based \hae detector with a dynamic data buffer on the client side, to detect the whole trace of an \hae, regardless of its length. As shown in Figure~\ref{fig:dynamic_data_buffer}, this data buffer aims to capture the entire high acceleration period, when the magnitude of linear acceleration is higher than the threshold (i.e. $5 m/s^2$). To cover more context before and after the period, $3$ seconds before the first threshold crossing point, and $3$ seconds after the last threshold crossing point are also included. The training dataset consists of short driving windows ($5$ seconds) around the peak of a \hae detected by phone sensors. \hbes are a special case of \hae with large deceleration on the vehicle's longitudinal moving direction.

Each \hae contains sensor data from the accelerometer, gyroscope, and \gps speed readings inside the detection window at $10 Hz$. The \haes from projection mode also include wheel speed readings from the vehicle. Table~\ref{tab:sensor_variables} lists feature names, units, shapes, and descriptions. To align the dimension and shape of sensor channels, we truncate sensor readings to $5$ seconds around the peak ($2.5$ seconds on each side) and linearly interpolate the data to $101$ samples. This transforms the shape of phone IMU sensor readings to $3$ channels by $101$ data points, and the shape of phone \gps speed and vehicle wheel speed variables to $1$ channel by $101$. As vehicle wheel speed is used to calculate labels for model training, a median filter with kernel size of $11$ is applied to the wheel speed time series to remove high-frequency noise.

\begin{table}[ht]
    \small
    \centering
    \begin{tabular}{c|c|c|c}\hline
        Name & Unit & Shape & Description\\\hline\hline
        Phone accelerometer~\cite{android-accelerometer} & $m/s^2$ & $(3, 101)$ & Acceleration force along 3 axes (including gravity)\\\hline
        Phone gyroscope~\cite{android-gyroscope} & $rad/s$ & $(3, 101)$ & Rate of rotation around 3 axes\\\hline
        Phone linear acceleration~\cite{android-linear-acceleration} & $m/s^2$ & $(3, 101)$ & Acceleration force along 3 axes (excluding gravity)\\\hline
        Phone \gps speed~\cite{android-gps-speed}& $m/s$ & $(1, 101)$ & Speed estimated from \gps location samples\\\hline
        Vehicle wheel speed~\cite{android-car-speed} & $m/s$ & $(1, 101)$ & Vehicle wheel speed from speedometer sensors\\\hline
    \end{tabular}
    \caption{Smartphone and vehicle sensor variables.}
    \label{tab:sensor_variables}
\end{table}

The vehicle’s longitudinal acceleration ($Accel_{v\_long}$) is calculated as the first derivative of the vehicle wheel speed ($Speed_v$). 

\begin{equation}
    Accel_{v\_long} = d(Speed_v)/dt
\end{equation}

A wheel speed-based \hbe is identified when the minimum  acceleration value on the longitudinal direction, $min(Accel_{v\_long})$, is lower than or equal to a predefined threshold. We empirically set this threshold value to $-5 m/s^2$, which is similar to values suggested by literature~\cite{simons2013trajectories, simons2012elevated, mao2021decision}. The  wheel speed-based \hbe, as described in the following formula, is treated as the ground truth label for the prediction model.

\begin{equation}
    label = (min(Accel_{v\_long} \leq -5))
\end{equation}

The phone dataset consists of smartphone sensor sequences from users who are actively navigating with Google Maps using their smartphones not in projection mode. We used this dataset to demonstrate the efficacy of \hbes as a safety proxy and to examine possible selection bias. In addition to raw phone sensor readings, the road segments where the \haes were sampled are also included. 

Table~\ref{tab:datasets_info} shows the detailed information about these two datasets during the same time period. Overall, the phone dataset contains $97\times$\haes, $22\times$distance, $31\times$duration, and $26\times$segment traversals compared to the projection mode dataset. Therefore, when applying the \hbe detection model trained on the projection mode dataset to the phone dataset, we can achieve greater than $20\times$coverage than the projection model dataset.

\begin{table}[ht]
    \small
    \centering
    \begin{tabular}{c|c|c|c|c}\hline
        Dataset type & Total \haes & Total Distance (\textsc{Km}) & Total Duration (H) & Segment Traversals\\\hline\hline
        Projection mode & $1.74\times10^8$ & $2.84\times10^9$ & $5.21\times10^8$ & $2.45\times10^{10}$\\\hline
        Phone mode & $1.68\times10^{10}$ & $6.26\times10^{10}$ & $1.63\times10^{10}$ & $6.37\times10^{11}$\\\hline
    \end{tabular}
    \caption{Information on the two datasets, including the number of total \haes, total distance in \textsc{Km}, duration in hours, and total road segment traversals.}
    \label{tab:datasets_info}
\end{table}

Lastly, we use collision datasets containing vehicle crash events and their corresponding locations. These datasets are usually collected and released to the public by local police or transportation departments regularly. In this work, we use collision data in five regions in US (the state of California~\cite{california-crash-data}, Chicago~\cite{chicago-crash-data}, Washington DC~\cite{dc-crash-data}, New York City~\cite{nyc-crash-data}, and Seattle~\cite{seattle-crash-data}) for evaluating the association between \hbes and crashes.

%% file: sections/methodology.tex
\section{Methodology}

Several existing studies have used smartphone sensors to measure vehicle dynamics~\cite{mohan2008nericell, thiagarajan2009vtrack, wang2015determining, chen2015invisible, liu2017bigroad}. For example, the accelerometer can be used to determine the longitudinal and lateral acceleration of a vehicle; the gyroscope reflects a vehicle’s rotation velocity while turning; and \gps signals can be used to estimate the speed of the vehicle. Detecting \hbes using smartphone sensors in real-life driving is difficult as smartphones may be located in any arbitrary location inside a vehicle, such as in a cup holder, a phone mount, or even the driver’s hands. Phone orientation with respect to the vehicle is unknown and may keep changing over time. Phone acceleration and gyroscope readings can be affected by movements unrelated to vehicle motions, such as phone usage by drivers or passengers. Simply applying a threshold-based heuristic to changes in raw phone accelerometer magnitude cannot distinguish kinematic signature from vehicle movement. \gps speed provides information about vehicle movement, but it also suffers from low sample rate and low accuracy, especially in urban areas.

To understand phone orientation inside a vehicle, most existing research either fixes the phone to a known orientation relative to the vehicle~\cite{hull2006cartel}, or uses coordinate alignment methods to estimate phone orientation inside the vehicle~\cite{wang2015determining, chen2015invisible, liu2017bigroad}. However, these methods cannot achieve high quality detection in large-scale, real-world deployments due to the complexity of phone usage and movement inside vehicles.

Machine learning has been used to deliver superior performance on complex tasks, such as computer vision, natural language processing, and signal processing. In particular, time series model architectures like \lstm~\cite{hochreiter1997long} and Transformer~\cite{vaswani2017attention} can be used to understand latent information underlying continuous sensor readings. However, training a machine learning model for \hbe detection requires large amounts of data with ground truth labels—which is generally not available in publicly accessible datasets. In this study we developed a system architecture to train \hbe prediction models based on smartphone sensors. 

\subsection{System Architecture}
\begin{figure}[ht]
    \centering
    \includegraphics[width=1.0\textwidth]{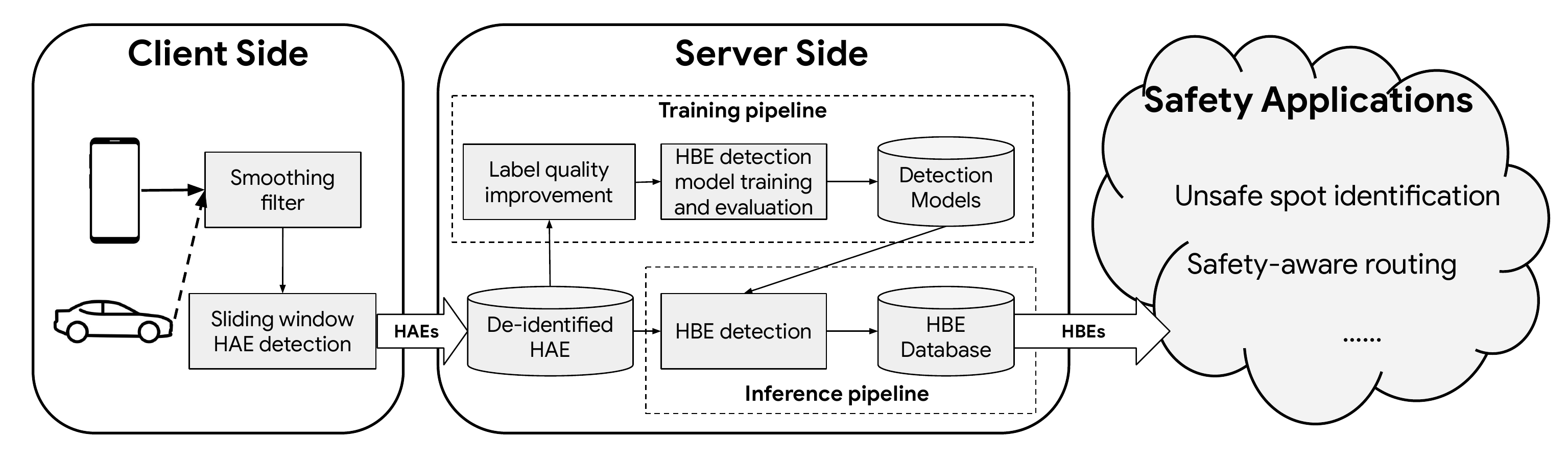}
    \caption{System architecture.}
    \label{fig:system_arch}
\end{figure}

We built a system to support \hbe detection model training and inference at large scale. As shown in Figure~\ref{fig:system_arch}, the system includes three main components: (1) Client side \hae detection, (2) Server side \hbe detection model training, and (3) Server side \hbe detection inference. The detected \hbes can be used for the subsequent safety applications, including unsafe spot identification, safety-aware routing, and safety impact measurement on specific features. 

On the client side, we implemented an \hae detection module on mobile phones. This module takes real-time sensor readings from the phone to identify \haes and upload the \haes to the cloud server. To remove high-frequency noise and reduce upload bandwidth consumption, sensor readings are passed to a smoothing filter. This filter averages data within each $100 ms$ window to smooth and downsample the high-frequency signals to $10 Hz$. The filtered data is passed to the sliding window \hae detection algorithm (introduced in Figure~\ref{fig:dynamic_data_buffer}), which emits an \hae if the maximum magnitude of the linear acceleration in the window is greater than a threshold (i.e. $5 m/s^2$). The detected \haes are uploaded to the server for \hbe detection model training and inference.

On the server side, the \hae dataset in projection mode serves as input for the training and inference pipelines. In the training pipeline, we construct a supervised learning model using \haes from the projection mode, which include features from the phone sensors and the binary labels, i.e., whether a wheel speed-based \hbe occurred. In the inference pipeline, the system applies the trained detection model to identify \hbes among the \haes. These identified \hbes are stored in a separate database for safety applications.

\subsection{Model Structure}

\begin{figure}[ht]
    \centering
    \includegraphics[width=0.5\textwidth]{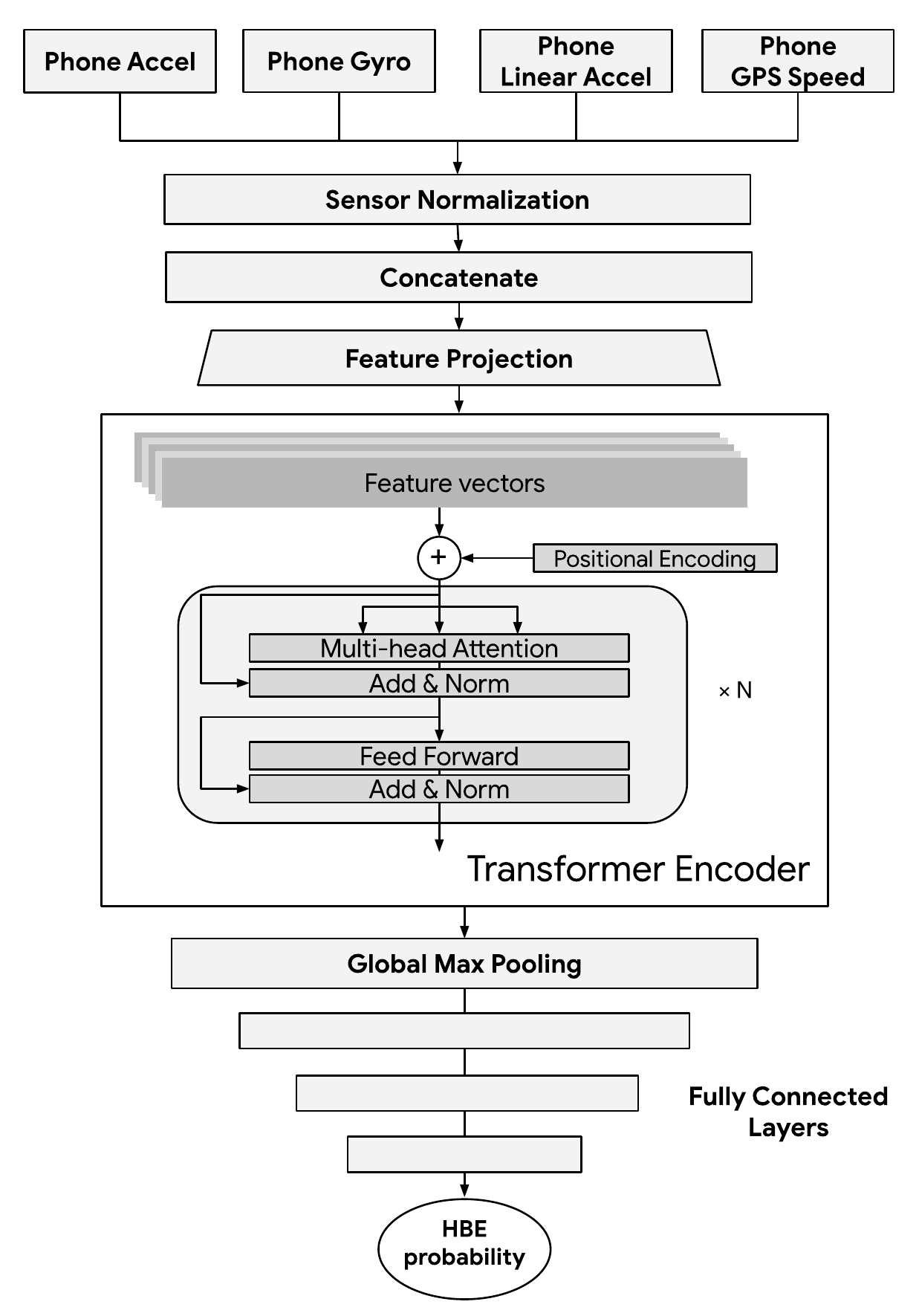}
    \caption{Model architecture.}
    \label{fig:model_arch}
\end{figure}

Figure~\ref{fig:model_arch} shows the Transformer-based model structure for \hbe detection. We first normalize sensor signals using precomputed mean and variance, followed by an early fusion approach to concatenate 3 channels of phone accelerometer, 3 channels of phone gyroscope, 3 channels of phone linear acceleration, and 1 channel of phone \gps speed together to construct a multi-channel time series tensor with a shape of (10, 101). A feature projection layer is used to expand feature dimension to (M, 101). After passing through a positional encoding layer, the tensor is then fed into a Transformer encoder to extract a latent representation. 

The Transformer~\cite{vaswani2017attention} is a deep learning model architecture that adopts the mechanism of self-attention, differentially weighting the significance of each part of the input data. It has been widely used in the field of natural language processing (NLP) and in computer vision. In this work, the Transformer encoder is composed of a stack of N Transformer layers, each is constructed by a multi-head self-attention unit and a feed-forward network. The latent representation passes through a global max pooling layer to concentrate to a (M, 1) sized feature vector, which is input into a classification head constructed with multiple fully connected layers. We use M=128 and N=6 in this model based on an extensive model architecture search. The final layer of the model uses a sigmoid activation function to output a probability between 0 and 1. As shown in Equation~\ref{equ:bce_loss}, the binary cross entropy between the output probability and the label is used as the loss function for model training. 

\begin{equation}
    loss = -\frac{1}{N}\sum_i^N y_i \log \hat{y_i} + (1-y_i) \log (1-\hat{y_i})
    \label{equ:bce_loss}
\end{equation}

\subsection{Alternative Models and Performance Evaluation Metrics}

Several state-of-the-art machine learning models and heuristics algorithms were compared with the Transformer models adopted in this study. The alternative machine learning models focus on different encoder structures, including fully connected neural network models, convolutional neural network (\cnn) models, and long short term memory models (\lstm) encoders. All alternative models have gone through a rigorous tuning process to find optimal performance as discussed in the following section.
 
In addition to the machine learning models, two heuristic detectors are also evaluated as baselines:
\begin{itemize}
    \item \textit{Accelerometer-based detector}: We used the negative of the maximum phone linear acceleration magnitude as the predicted vehicle deceleration of the event. With a predefined threshold, we can build a binary classification model that predicts an event as an \hbe if its predicted vehicle deceleration is below this threshold. By changing the threshold, the recall and precision of this model can change. This heuristic detector does not attempt to correct the coordinate misalignment between phone and vehicle, and assumes all linear accelerations are caused by vehicle deceleration. 
    \item \textit{\gps speed-based detector}: We estimate vehicle longitudinal acceleration using smartphone \gps speed readings, using its minimum value as the predicted vehicle deceleration. Similarly, we can change the detection threshold to measure the precision and recall trade-off of this model.
\end{itemize}

We evaluate the model performance using the precision-recall (\pr) and receiver operating characteristic (\roc) curves, and their corresponding area under the curve (\auc) scores. The \pr curve shows the precision and recall tradeoff under the real data distribution, and the \roc curve shows the relationship between the true positive rate (\tpr) and the false positive rate (\fpr)  under different classification thresholds. In addition, we also evaluate the precision of the model under certain recall thresholds, and discuss the potential use cases of these different models.

\subsection{Model Training and Comparison}
We train the model using \haes from the projection mode dataset. We use the binary cross-entropy loss and Adam optimizer to train the model. A comprehensive hyperparameter search on learning rate, epsilon, batch size, model structure (including number of layers, units/kernals in each layers, etc.) is conducted for each model. 

\textbf{Encoder architectures.} We experimented with a number of model architectures to find the best performing model. One search space is to replace the Transformer encoder in the model with other commonly used architectures, including fully connected (\fc-based) encoders, convolutional neural network (\cnn)-based encoders, and long-short-term-memory (\lstm)-based encoders. Table~\ref{tab:encoder_arch} shows four different models from each encoder type using similar numbers of model parameters. The results show that the Transformer-based encoders attain the best \prauc compared to other encoders. \lstm-based encoders and \cnn-based encoders attain very similar \prauc scores, and \fc has the poorest performance overall. This suggests that models incorporating time-series characteristics, like Transformer, \lstm, and \cnn, can be very powerful tools for detecting \hbes from phone sensor readings.

\begin{table}[ht]
    \centering
    \begin{tabular}{c|c|c}\hline
        Encoder Type & Parameters & \prauc\\\hline\hline
        \fc & 380,321 & 0.795\\\hline
        \cnn & 403,137 & 0.831\\\hline
        \lstm & 342,421 & 0.831\\\hline
        Transformer & 268,501 & 0.833\\\hline
    \end{tabular}
    \caption{Area under the precision-recall curve (\prauc) of four model encoder architectures.}
    \label{tab:encoder_arch}
\end{table}

\textbf{Feature space.} We also performed an ablation study on all combinations of the four features. For each feature, inclusion as a model input was optional. We evaluated the resulting 15 distinct feature combinations under the same Transformer model structure, dataset, and training configurations. Table~\ref{tab:model_performance_single_feature} shows the 4 combinations with only one feature. The model using phone accelerometer input has the highest \prauc score (0.745). The models using linear acceleration and \gps speed attain a 0.645 and 0.624 \prauc respectively, also showing reasonable predictive power. The gyroscope model provides the least prower with 0.346 \prauc. Table~\ref{tab:model_performance_multiple_feature} shows the top four feature combinations and the corresponding \praucs. The feature combination leveraging all 4 features achieves the best \prauc score. All top-performing combinations contain phone accelerometer and \gps speed. 

\begin{table}[ht]
    \parbox{.40\linewidth}{
        \centering
        \begin{tabular}{c|c}\hline
            Features & \prauc\\\hline\hline
            Accelerometer & 0.745\\\hline
            Gyroscope & 0.346\\\hline
            Linear Acceleration & 0.645\\\hline
            \gps Speed & 0.624\\\hline
        \end{tabular}
        \caption{Transformer model performance using a single feature.}
        \label{tab:model_performance_single_feature}
    }
    \hfill
    \parbox{.55\linewidth}{
        \centering
        \begin{tabular}{c|c}\hline
            Feature combinations & \prauc\\\hline\hline
            Accel + Gyro + Linear Accel + Speed & 0.833\\\hline
            Accel + Gyro + Speed & 0.829\\\hline
            Accel + Linear Accel + Speed & 0.818\\\hline
            Accel + Speed & 0.814\\\hline
        \end{tabular}
        \caption{Transformer model performance using multiple features.}
        \label{tab:model_performance_multiple_feature}
    }
\end{table}

\begin{figure}[ht]
    \centering
    \includegraphics[width=.45\textwidth]{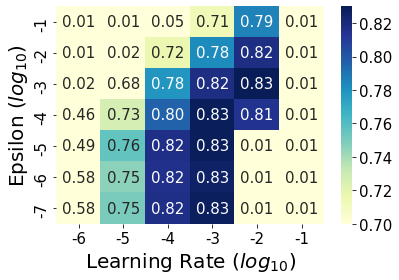}
    \caption{\prauc of models trained with different values of learning rate and epsilon in the Adam optimizer.}
    \label{fig:hyperparameter_sweep}
\end{figure}

\textbf{Hyperparameter sweep.} We further conduct a hyperparameter sweep on learning rate and epsilon in the Adam optimizer. Figure~\ref{fig:hyperparameter_sweep} shows the final model \prauc score trained using different values for the model hyperparameters learning rate and epsilon. The labels on the axis indicate the learning rate and epsilon in $\log_{10}$ scale. We found that $10^{-3}$ is the best learning rate among the values explored, and smaller epsilon (e.g. $10^{-4}-10^{-7}$) works better than larger epsilon values in our search space.

%% file: sections/result.tex
\section{Results}

\subsection{Model Accuracy Evaluation}

\begin{figure}[ht]
\centering
\begin{minipage}{.48\textwidth}
  \centering
  \includegraphics[width=0.9\linewidth]{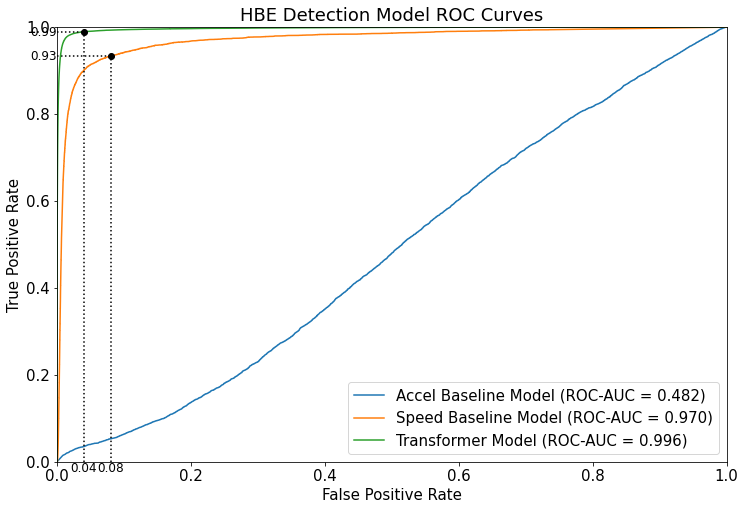}
  \caption{\rocauc of \hbe detection model.}
  \label{fig:roc_auc}
\end{minipage}
\begin{minipage}{.48\textwidth}
  \centering
  \includegraphics[width=0.9\linewidth]{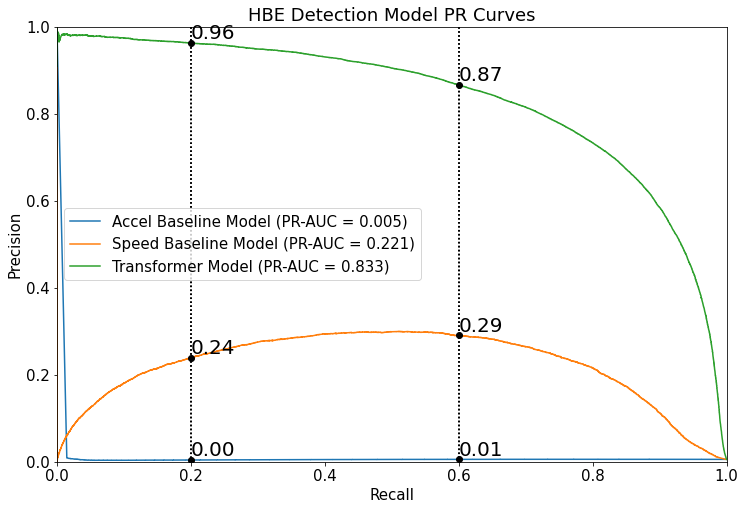}
  \caption{\prauc of \hbe detection model.}
  \label{fig:pr_auc}
\end{minipage}
\end{figure}

The \roc curves and corresponding \rocauc scores of two heuristic detectors and the Transformer-based model are shown in Figure~\ref{fig:roc_auc}. To generate \roc curves for the two heuristic detectors, we apply different threshold values on the predicted vehicle decelerations, and measure \tpr and \fpr at each threshold. As the accelerometer heuristic detector creates too many false positive predictions, its \rocauc score is only 0.482, which is worse than random guessing. The \gps speed-based heuristic detector achieves a 0.970 \rocauc score, indicating reasonable predictive power of \gps speed for \hbe detection. The Transformer-based \hbe detection model substantially outperforms the heuristics models with a 0.996 \rocauc score. When thresholding on a balanced trade-off point, the Transformer model can achieve 0.99 \tpr with only 0.04 \fpr. 

To further understand how the model works with the real data distribution, we evaluate the precision-recall (\pr) curves and corresponding \prauc scores of three models as shown in Figure~\ref{fig:pr_auc}. To generate \pr curves for the two heuristic detectors, we apply different thresholding values on the predicted vehicle decelerations, and measure precision and recall pairs. 

As can be seen, the accelerometer-based heuristic approach performs poorly with a 0.005 \prauc score. This is likely due to large amounts of phone movement unrelated to hard-braking. While the \gps speed-based detector has a decent \rocauc, it only achieves a 0.221 \prauc. This is likely due to severely imbalanced data, i.e. substantially more non-\hbes than \hbes in the \hae dataset. The \gps speed heuristic detector does not have a good precision and recall tradeoff since  precision does not increase monotonically as recall decreases. Compared with two heuristic detectors, the Transformer-based \hbe detector achieves the best \prauc score of 0.833, and has a better precision and recall tradeoff. We find that the Transformer-based detector achieves 0.96 precision at 0.2 recall, and 0.87 precision at 0.6 recall. This provides flexibility to choose the probability threshold based on the precision and recall requirement of different applications – higher thresholds would provide more precise results but reduce coverage. 

\subsection{Evaluating Aggregate HBE Data as a Safety Surrogate}
A primary objective of \hbe detection is to infer traffic safety, which has been traditionally measured by crashes~\cite{highway-safety-manual}. A sufficient surrogate safety proxy needs to have robust spatial correlations with collision data. In this section, we show our aggregations of inferred phone-only \hbes achieve high correlation with collision data, with better correlation compared with aggregations of phone-only \haes. This correlation increases as we use a higher precision model in most areas.

To conduct this experiment, we pass \haes from the phone dataset across a multi-day period through the Transformer-based \hbe detection model to generate a probability of being an \hbe for each \hae. We then aggregate \hbes by road segment level. The estimated \hbe rate for a road segment is the quotient of the \hbe count and the sum of the distance driven on that segment. A probability threshold on the Transformer model outputs is needed to determine whether a specific \hae is an \hbe or not. Using a higher probability threshold means the model can output \hbes with higher precision but lower recall. On the other extreme, using zero as the threshold makes the \hbe rate equal to \hae rate, since no \haes are filtered out. Multiple threshold values are evaluated to avoid subjectivity in selecting just one value. 

In addition to the \hbe rate, we also map collisions from publicly available crash datasets in five regions, including the state of California~\cite{california-crash-data}, Chicago~\cite{chicago-crash-data}, Washington DC~\cite{dc-crash-data}, New York City~\cite{nyc-crash-data}, and Seattle~\cite{seattle-crash-data}, to these road segments. The collision rate for a segment is calculated as the quotient of the collision count and the sum of the distance driven on that segment with Google Maps navigation. Under the approximation that the fraction of traffic using Google Maps navigation is roughly constant, this collision rate would be approximately proportional to the actual collision rate. Correlation is invariant to changes in scale.

\begin{figure}[ht]
    \centering
    \includegraphics[width=.6\textwidth]{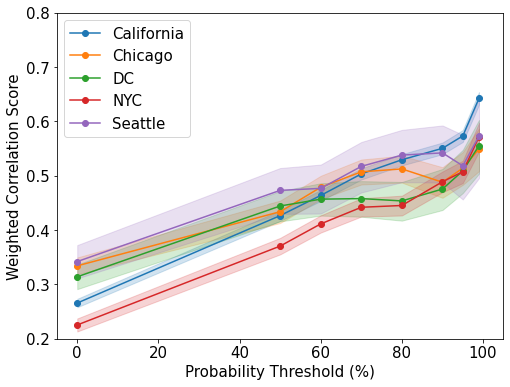}
    \caption{Weighted correlation between inferred phone \hbe rate and collision rate at segment level in five regions.}
    \label{fig:safety_proxy_eval}
\end{figure}

In these five regions with public collision data available, we calculate the weighted Pearson correlation~\cite{pearson-correlation-coefficient} coefficient between collision rate and \hbe rate with a variety of \hbe probability thresholds (0, 0.5, 0.6, 0.7, 0.8, 0.9, 0.95, 0.99). In calculating correlation, each road segment is one point (\hbe rate, collision rate), and its weight is the total distance traveled, which is also the denominator for collision rate. If a segment has only a little distance driven on it and happens to have some \hbes or collisions, its observed rate can be misleadingly high. Simple correlation can be influenced by such points, to either exaggerate or suppress the correlation depending on whether or not those points fall along the main trend. The observed rate for a segment with a large distance driven on it has much less uncertainty — there have been many more opportunities to have, or not have, an \hbe or collision. By using weighted correlation we account for the inverse relationship between the rate variance and the observed distance driven.

Figure~\ref{fig:safety_proxy_eval} shows the correlation results under segment-level aggregation. We can clearly see that, in most cases, the correlation score has an increasing trend as we increase the probability threshold. This demonstrates that our \hbe detection model is capable of filtering out noisy \haes and helping us build a better safety surrogate for road collision likelihood.

\subsection{Model Fairness Evaluation}
As part of Google's AI principles~\cite{google-ai-principles}, we seek to avoid unintended, unjust impact on people, particularly those related to sensitive characteristics. In line with these principles, we aim to make a model that performs equally well for all users, regardless of sensitive characteristics. To this end, we studied the accuracy of the \hbe detection models for various region-based socioeconomic groups. For example, we assessed the effect on regions having different median income levels and different levels of educational attainment. We aim for Equality of Opportunity and Equalized Odds in which model quality is roughly equivalent across selected subgroups~\cite{hardt2016equality}.

In this study, we mapped \haes in projection mode in the US to corresponding zip codes and joined the data with a US census report~\cite{us-census-data} which includes median income and level of educational attainment rates (bachelor’s or higher degree). We then aggregated the \hbes into various income and educational attainment groups before evaluating the accuracy of trained \hbe detection models within each subgroup. Table~\ref{tab:fairness_median_income} and~\ref{tab:fairness_education} summarize the results.

Table~\ref{tab:fairness_median_income} and~\ref{tab:fairness_education} show \prauc scores of the model according to various socioeconomic  groups and the relative percentage difference compared to the best group (in bold). Results in the first row are generated using real data distribution, in which the number of positive samples (\hbes) are significantly lower than the number of negative samples (non-\hbes). The largest performance gaps in \prauc scores between higher and lower income/education groups are $3.27\%$ and $5.01\%$ correspondingly. We found that this gap is mainly due to the different degree of data skew between groups. Lower income or education groups tend to have lower positive sample rates, which can lower the \prauc score, even if the model has a similar false positive rate across groups. 

We use the bane of skew method~\cite{lampert2014bane} to fix the inconsistent data skew between groups. As shown in Equation~\ref{equ:bane_of_skew}, a precision value $P_\pi(\theta)$ under one data skew $\pi$ can be transformed into a precision value $P_{\pi'}(\theta)$ under another skew $\pi'$, where the data skew $\pi$ is defined as the ratio of positive samples in the dataset.

\begin{equation}
    P_{\pi'}(\theta) = \frac{\pi'}{\pi'+(1-\pi')\frac{\pi}{1-\pi}[\frac{1}{P_\pi(\theta)}-1]}
    \label{equ:bane_of_skew}
\end{equation}

\begin{table}[ht]
    \centering
    \small
    \begin{tabular}{c|c|c|c|c|c}\hline
        \makecell{Median\\ Income} & $<50K$ & $50K-100K$ & $100K-150K$ & $150K-200K$ & $>200K$\\\hline\hline
        \prauc & \textbf{0.900} & $0.871 (\textcolor{red}{-3.27\%})$ & $0.875 (-2.87\%)$ & $0.876 (-2.77\%)$ & $0.890 (-1.13\%)$\\\hline
        \makecell{\prauc \\(fixed skew)} & \textbf{0.9994} & $0.9991 (\textcolor{red}{-0.030\%})$ & $0.9993 (-0.008\%)$ & $0.9992 (-0.022\%)$ & $0.9992 (-0.022\%)$\\\hline
    \end{tabular}
    \caption{Model performance by median income.}
    \label{tab:fairness_median_income}
\end{table}

\begin{table}[ht]
    \centering
    \small
    \begin{tabular}{c|c|c|c|c|c}\hline
        \makecell{Educational \\Attainment} & $<20\%$ & $20-40\%$ & $40-60\%$ & $60-80\%$ & $80-100\%$\\\hline\hline
        \prauc & $0.839	(\textcolor{red}{-5.01\%})$ & \textbf{0.884} & $0.879 (-0.53\%)$ & $0.872 (-1.34\%)$ & $0.877 (-0.79\%)$\\\hline
        \makecell{\prauc \\(fixed skew)} & $0.9988 (\textcolor{red}{-0.061\%})$ & \textbf{0.9994} & $0.9993 (-0.010\%)$ & $0.9992 (-0.020\%)$ & $0.9993 (-0.013\%)$\\\hline
    \end{tabular}
    \caption{Model performance by educational attainment: bachelor or higher degree rate.}
    \label{tab:fairness_education}
\end{table}

We use this method to transform the precision values in each group from its original skew to $\frac{1}{2}$ — which is the skew for a balanced dataset (having the same amount of positive and negative samples). The results in the second row of each table show the \prauc score after we fix the data skew. The largest performance gaps in \prauc scores become 0.030\% and 0.061\% between income and education groups. These results demonstrate that the model performance is quite similar across different socioeconomic groups.

\subsection{Selection Bias Evaluation}
Training a model using data from projection mode and applying it to phone-only circumstances can greatly increase the coverage rate for road segments ($26\times$ according to Table~\ref{tab:datasets_info}), but does raise concerns about potential selection bias. 

\begin{figure}[ht]
    \centering
    \includegraphics[width=.6\textwidth]{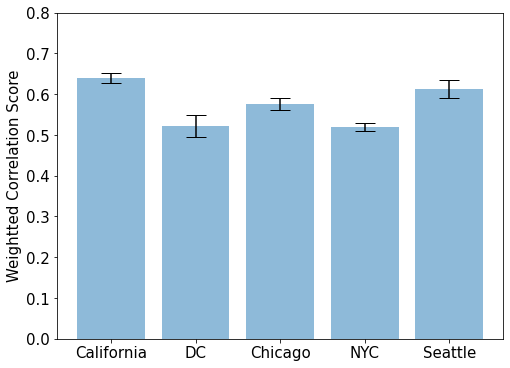}
    \caption{High corrlation between the phone inferred \hbe rate and wheel-speed \hbe rate at the segment level.}
    \label{fig:selection_bias}
\end{figure}

To evaluate this, we calculate segment-level correlation scores between projection-mode \hbe rate and phone inferred \hbe rate in five regions similar to the Safety Surrogate evaluation. In this analysis, the projection-mode \hbe rate of each road segment is calculated by dividing the number of projection-mode \hbes by the collective distance of projection mode traversals of that segment. Phone inferred \hbe rate is calculated using phone inferred \hbes detected with $0.5$ probability threshold as the numerator, and distance from phone-only segment traversals as the denominator on the same segment. Figure~\ref{fig:selection_bias} shows high correlation scores $(>0.5)$ with very tight $95\%$ confidence intervals in all five regions. The strong correlation scores show the distribution of the phone inferred \hbes and wheel-speed \hbes are similar and no meaningful bias observed.  

%% file: sections/conclusion.tex
\section{Summary and Conclusions}
Traffic collisions lead to millions of lives lost and a tremendous amount of property damage globally each year. Traditionally, crashes are the primary measure of risk, but they are rare for specific road segments. Also, a lack of consistent sources and data collection methods brings challenges in risk assessment and prediction. The ubiquitous availability of mobile mapping services on smartphones provides a scalable and low cost source to assess driving risk at network level.  Constantly connected smartphones also enable safety intervention to drivers directly through safe routing and notifications.   

This paper presents a novel machine learning approach to detect hard-braking events using smartphone sensors. By not relying on vehicle sensors, \hbe detection can expand the coverage to billions of Google Map users, two orders of magnitude higher than coverage provided by Android Auto vehicle sensors. We evaluated multiple state-of-the-art ML architectures, including fully connected neural network, \cnn, \lstm, and Transformer-based models, and found that Transformer-based models have the best performance for detecting \hbes. Our detection model achieved a 0.83 \prauc score, which is $3.8\times$ better than a phone \gps speed-based heuristic model, and $166.6\times$ better than a phone accelerometer-based heuristic model.

To confirm that the detected \hbes do represent real crash risk, we evaluated the correlation between detected \hbes and collision data from several cities. The results show a strong correlation with traffic collisions at the road segment-level. The strong correlation with crash risk validates the use of \hbes as a crash surrogate in safety applications. We aim to avoid unjust impacts on people, particularly those related to sensitive characteristics, and thus conducted a fairness evaluation to ensure that the safety benefits of our system would be shared equally across selected subgroups of users. We conducted a selection bias evaluation that showed that the distribution of detected \hbes and ground truth wheel-speed \hbes are similar and thus no meaningful selection bias was observed.  

The availability of phone-based \hbe detection enables a variety of safety applications and services at unprecedented scale, globally. For example, continuously tracking \hbes in aggregate level can help us identify high risk spots at road network level thanks to significant data coverage. The high density phone-based \hbe data also enables proactive safety countermeasures, for example in safety-aware routing platforms.  This can be achieved by including historical or real-time \hbe rates on given road segments in the cost function used in  navigation routing algorithms. As discussed in a blog post~\cite{safe-routing-blog}, a safety-aware routing platform could help reduce collision risk, while also providing drivers with safer and more comfortable routes.

The primary goal of this work is to develop an \hbe detection model that is accurate enough to serve as a useful safety proxy. Consequently, the study did not focus on identifying the most efficient model architecture that would preserve model accuracy while reducing parameter size. A future research topic is to improve the model efficiency, reduce computation consumption, and potentially enable client-side inference. 

The high prevalence of smartphones provides a low-cost means to collect driving kinematics data at driver population levels. To elicit metrics that can infer traffic safety will enable active safety countermeasures such as safety-aware routing. The methodology developed in this paper allows identification of hard-braking events for the entire road network and provides a novel data source for traffic safety improvement efforts. 

%% file: sections/acknowledgement.tex
\section*{Acknowledgement}
The authors thank Tim Long, Ting-You Wang, Divya Jetley, Brian Ferris, Greg Phipps, Geoff Clark, Aldi Fahrezi, Jonathan Waltz, Jon Kimbel, Eric Dickinson, Genevieve Park, Kevin Eustice, Marco Gruteser, Eric Tholomé, and Blaise Ag\"uera y Arcas for their inspiration, contribution and guidance to this work.